\pdfoutput=1

\documentclass[11pt]{article}

\usepackage[]{ACL2023}

\usepackage{times}
\usepackage{latexsym}

\usepackage[T1]{fontenc}

\usepackage[utf8]{inputenc}

\usepackage{microtype}

\usepackage{multirow}
\usepackage{graphicx}  
\usepackage{caption}
\usepackage{booktabs}
\usepackage{enumitem}
\usepackage{subcaption}
\usepackage{latexsym}
\usepackage{soul}
\usepackage{microtype}
\usepackage{inconsolata}
\usepackage{microtype}
\usepackage{todonotes}
\usepackage{tabularx}
\usepackage[]{graphicx}
\usepackage{enumitem}

\title{DisCGen: A Framework for Discourse-Informed Counterspeech Generation} 

\author{Sabit Hassan \\ School of Computing and Information \\  University of Pittsburgh, Pittsburgh, PA\\
\texttt{sabit.hassan@pitt.edu}
        \And
        Malihe Alikhani \\ Khoury College of Computer Science \\ Northeastern University, Boston, MA\\
        \texttt{m.alikhani@northeastern.edu}
        }

\begin{document}
\maketitle
\begin{abstract}
Counterspeech can be an effective
method for battling hateful content on social media. Automated counterspeech generation can aid in this process. Generated counterspeech, however, can be viable only when grounded in the context of topic, audience and sensitivity as these factors influence both the efficacy and appropriateness. In this work, we propose a novel framework based on theories of discourse to study the inferential links that connect counter speeches to the hateful comment. Within this framework, we propose: i) a taxonomy of counterspeech derived from discourse frameworks, and ii) discourse-informed prompting strategies for generating contextually-grounded counterspeech. To construct and validate this framework, we present a process for collecting an in-the-wild dataset of counterspeech from Reddit. Using this process, we manually annotate a dataset of \textbf{3.9k} Reddit comment pairs for the presence of hatespeech and counterspeech\footnote{Our code and data can be requested from here: https://github.com/sabithsn/DisCGen}.
The positive pairs are annotated for 10 classes in our proposed taxonomy. We annotate these pairs with paraphrased counterparts to remove offensiveness and first-person references. We show that by using our dataset and framework, large language models can generate contextually-grounded counterspeech informed by theories of discourse. According to our human evaluation, our approaches can act as a safeguard against critical failures of discourse-agnostic models. 
\end{abstract}

\section{Introduction}
A promising countermeasure to hatespeech is \textit{counterspeech} \cite{Mathew2018ThouSN} \textemdash any response that counters hateful and offensive content, at times referred to as \textit{Counter-Narrative} \cite{fanton-etal-2021-human}. Counterspeech do not appear in isolation, but as an integral component of a broader discourse. In this work, we leverage theories of discourse \cite{Asher2005LogicsOC} to capture the context that the counterspeech appears in. 

The interpretation of a counterspeech is shaped by its relevance to the topic, its intended audience, and its sensitivity to the matter. For instance, a counterspeech posing a \textit{Probing Question} might resonate differently with an audience compared to one providing a \textit{Correction}. We propose a discourse-aware framework, \textbf{DisCGen}, to study these different types of counter speeches and how we can potentially generate them automatically.
\textbf{DisCGen} consists of a taxonomy of counterspeech based on discourse relations and discourse-augmented prompting strategies. The taxonomy is derived from \textbf{Segmented Discourse Representation Theory (SDRT)} \cite{Asher2005LogicsOC}. 

Since there is no existing counterspeech dataset with discourse relations, we construct the first dataset from Reddit to construct and validate our framework. We choose Reddit as the data source as in-the-wild data from Reddit is likely to have more diversity compared to Nichesourced or Crowdsourced datasets that contain counterspeech written by NGO workers \cite{chung-etal-2019-conan} or Mechanical Turk annotators \cite{qian-etal-2019-benchmark}. Diversity in the dataset is important to demonstrate the flexibility of our framework.

Constructing an in-the-wild dataset, however, is challenging as the percentage of comments forming hatespeech-counterspeech pairs is very small on Reddit. As such, we follow a two-stage process for collecting \textit{effective} counterspeech in-the-wild. We manually annotate a dataset \textbf{3.9K} Reddit comment pairs for the presence of hatespeech-counterspeech pairs. We manually annotate \textbf{250} positive pairs of hatespeech-counterspeech with SDRT relations. We paraphrase the counterspeech manually to remove profanity and first-person references while retaining the original content and linguistic style. We also annotate the positive samples with the \textbf{targeted group} in the hateful comment. While the full dataset can be used for identifying effective counterspeech, the positive pairs can be used with our framework for counterspeech generation.

Lastly, we combine the proposed discourse-based taxonomy with prompting strategies in our framework. We compare Large Language Models (LLMs) under different settings. In the first setting, discourse relations are provided for both the examples in the prompt and for the inference text. In the second scenario, discourse relations are provided only for the prompt examples. These models are compared with a baseline discourse agnostic approach. 
Our analysis shows greater diversity in discourse relations preserved in the discourse-informed approach compared to discourse-agnostic prompting. We show that, in both discourse-informed settings, LLMs are capable of generating highly accurate counterspeeches ( \textbf{>95\%} cases) and also respect corresponding discourse relations (\textbf{74\%} and \textbf{90\%} for the two strategies respectively). Further, our human evaluation shows that our strategies can act as safeguard against critical failures that discourse-agnostic LLMs are susceptible to. Thus, the contributions of this paper are:

\begin{itemize}[leftmargin=*]
    \item A novel framework for \textbf{discourse-informed} counterspeech generation that comprises of: i) a discourse-based taxonomy of counterspeech and ii) discourse-informed prompting strategies.
    \item A process for collecting an \textbf{in-the-wild} dataset of effective counterspeech from Reddit.
    \item First dataset of \textbf{3.9K} pairs of Reddit comments annotated for hatespeech-counterspeech, with \textbf{250} positive pairs annotated for: i) taxonomy derived from \textbf{SDRT}, ii) \textbf{paraphrasing} removing offensiveness and first-person references, and iii) \textbf{target group} that the hateful comment attacks.
    
\end{itemize}

\section {Related Work}
While offensive content on social media has gained much recent interest \cite{ye-etal-2023-multilingual}, work on counterspeech is still under-explored. \citet{Benesch2016counterspeech} conduct a field study of counterspeech on Twitter and list eight associated strategies from a social angle. The nichesourced CONAN \cite{chung-etal-2019-conan} dataset and its subsequent variations 
\cite{fanton-etal-2021-human,Bonaldi2022HumanMachineCA}, 
contain counterspeech written by NGO workers. \citet{qian-etal-2019-benchmark} inject counterspeech written by Mechanical Turk workers into conversations from Reddit and Gab. Our dataset is the first to contain counterspeech collected directly from Reddit, written by Reddit users. \citet{Mathew2018AnalyzingTH} and \citet{Mathew2018ThouSN} collect counterspeech from Twitter and YouTube respectively. While these are written by social media users, they are not suitable for \textit{generation} models due to presence of short and offensive counters. Our data is specifically curated for generation with removal of profanity and first-person references. Different from the existing works, ours is the first to present a discourse-based taxonomy and a dataset annotated with discourse relations.

A few recent works have studied counterspeech generation. \citet{bonaldi2023weigh} propose an attention based regularization with GPT-2 to generate more specific counter narratives. \citet{zhu-bhat-2021-generate} first generate multiple candidates, filter out ungrammatical ones, and then selects the most relevant countersppech. \citet{chung-etal-2021-towards} study generation of knowledge-grounded counternarratives using an external database. \citet{ashida-komachi-2022-towards} show the effectiveness of prompting large language models (LLMs) for generating counterspeech. \citet{vallecillo2023automatic} show that GPT-3 is more capable of generating counter narratives compared to other large language models. Ours is the first work to provide a framework for discourse-aware counterspeech generation.

Discourse relations have been proposed as a mechanism for controlling generation, shown to aid summarization \cite{Cohan2018ADA,Xu2020DiscourseAwareNE}, style transfer \cite{atwell-etal-2022-appdia}, and question answering \cite{huang2021dagn, xu2022we}. \citet{bosselut-etal-2018-discourse} show that discourse-aware models can generate more coherent texts. None of the aforementioned works however, target counterspeech generation. To our knowledge, our work is also the first to integrate discourse-based framework within prompting strategies for LLMs.

\section{Framework}
Our framework consists of a discourse-based taxonomy and two prompting strategies.
\subsection{Discourse-based Taxonomy} 
The style and efficacy of counterspeech is often dictated by its context, which is typically offensive or hateful content. Thus, we aim to identify different types of counterspeech through the lens of discourse relations. For the remainder of the paper, we use hatespeech as an umbrella term for offensive/hateful content.

With the help of linguist annotators, we explored ways different discourse theories such as  Penn Discourse Treebank (PDTB) \cite{Prasad2008ThePD}, Rhetorical Structure Theory (RST) \cite{Mann1988RhetoricalST} and Segmented Discourse Representation Theory (SDRT) \cite{Asher2005LogicsOC} might help us study the inferential links that connect hateful comments with counterspeech. Our initial investigation informed us that SDRT labels could closely model these inferential links. Thus, we decide to choose SDRT as primary source for our taxonomy.

We annotate 250 counterspeeches (Section \ref{dataset}) to construct a  discourse-based taxonomy of counterspeech that contain 10 classes adapted from SDRT relations defined in \citet{asher-etal-2016-discourse}:

\begin{itemize}[leftmargin=*]
\item \textit{Acknowledgment} when the counterspeech signals an understanding. While \citet{asher-etal-2016-discourse} include both understanding and acceptance as acknowledgment, acceptance is not considered in our definition as counterspeech should not agree with the hatespeech. 


\item \textit{Clarification Question} when the counterspeech asks questions to clarify information presented in the hatespeech, analogous to \citet{asher-etal-2016-discourse}.

\item \textit{Comment} when the counterspeech provides an opinion or evaluation of the content
in hatespeech, analogous to \citet{asher-etal-2016-discourse}.

\item \textit{Correction} when the counterspeech corrects an argument/fact presented in the hatespeech, analogous to \citet{asher-etal-2016-discourse}. 

\item \textit{Contrast} when the counterspeech provides a contrasting argument to hatespeech, analogous to \citet{asher-etal-2016-discourse}. 

\item \textit{Elaboration} when the counterspeech expands on the scenario presented in the hatespeech. Differing from \citet{asher-etal-2016-discourse}, counterspeech does not elaborate on its own argument, but offers a broader perspective on the hatespeech.

\item \textit{Probing question}, when the counterspeech asks a question intending to acquire more information, similar to Q-Elab in \citet{asher-etal-2016-discourse}. 

\item \textit{Explanation} when the counterspeech offers an explanation of a situation presented in the hatespeech, similar to \citet{asher-etal-2016-discourse}.

\item \textit{Parallel} when the counterspeech shows commonality between hatespeech and an external scenario, a special case of \citet{asher-etal-2016-discourse}.

\item \textit{Result} when the counterspeech connects the consequences to the content of hatespeech. The consequence is a special case of "effect" in \citet{asher-etal-2016-discourse}.

\end{itemize}

\subsection{Prompting Scenarios}
\label{prompt-strats}
We propose prompting strategies for two scenarios that use discourse relations in our taxonomy.\\

\noindent\textbf{Strategy 1:} The LLM is provided with discourse relations only for prompt examples. The LLM is asked to decide an appropriate discourse relation for inference text, and then generate a counterspeech .\\
\textbf{Strategy 2:} The LLM is provided with discourse relations for prompt examples and also inference text. The LLM is then asked to generate counterspeech respecting discourse relation of the inference text.\\ 

Strategy 1 is to be applied when there is no prior information about the type of counterspeech that should be generated. The model learns from prompting examples the type of discourse relations it should maintain with respect to context. Strategy 2 is to be applied when the desired discourse relation is known beforehand.



\begin{figure*}
    \centering
    \includegraphics[width=0.82\linewidth]{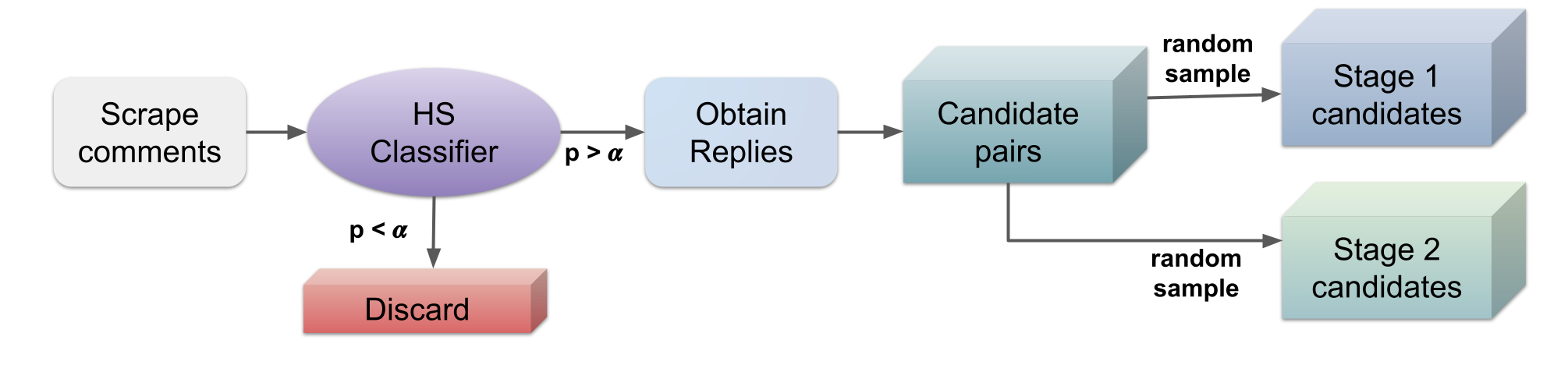}
    \caption{Our data collection pipeline. Comments and their replies are scraped from Reddit, and then run through a hatespeech classifier. If the classifier confidence falls below threshold $\alpha$ then they are discarded, else their replies are obtained to form pairs. These pairs are randomly split into two buckets for two stages of annotation. }
    \label{fig:method}
\end{figure*}
\begin{figure*}
    \centering
    \includegraphics[width=0.85\linewidth]{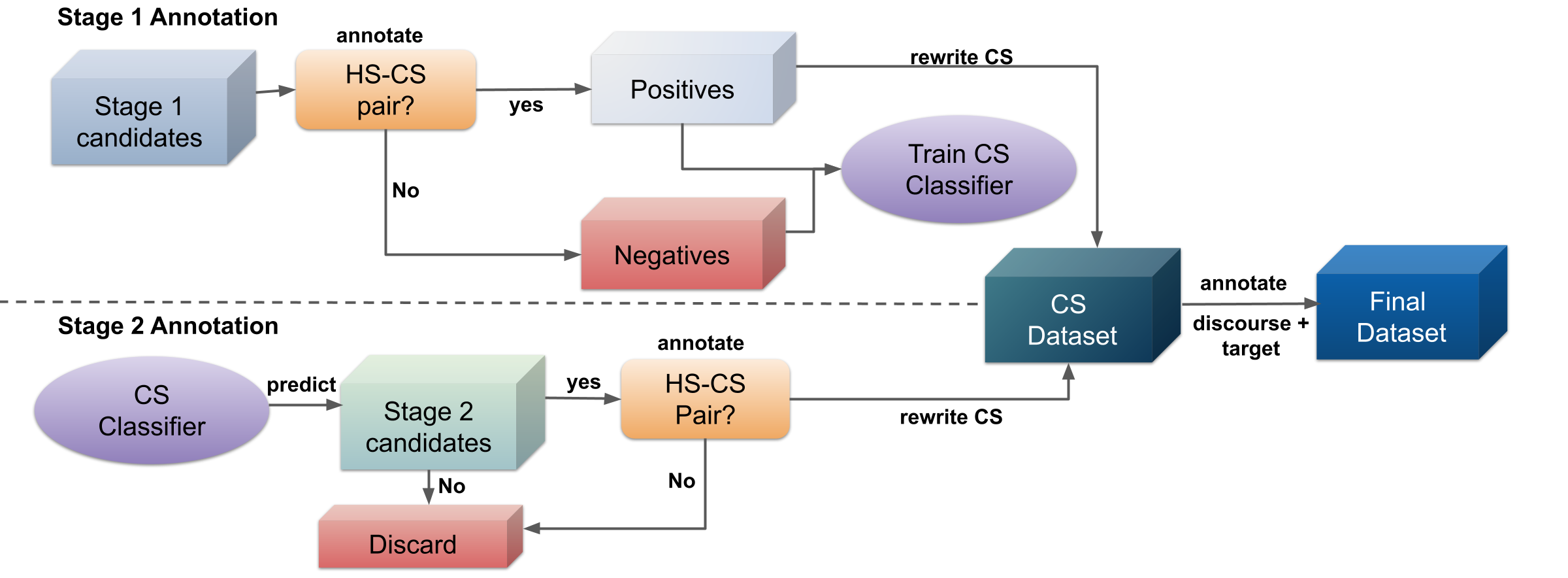}
    \caption{Our process for constructing in-the-wild dataset from Reddit. First set of random samples are annotated for hatespeech-counterspeech. Then we train a classifier to predict counterspeech, and then perform a final round of annotation. We rewrite the counterspeech if it is offensive or first-person. Finally, we annotate for discourse relations and target groups. In the figure, HS refers to hatespeech and CS refers to counterspeech.}
    \label{fig:method}
\end{figure*}
\section {Dataset}
\label{dataset}
In this section, we first outline our data collection pipeline. Then, we describe our two-stage process for constructing our dataset, followed by annotation protocol and inter-annotator agreement. Lastly, we analyze distributions in our dataset\footnote{Our data collection was approved by our institution's ethics review board (anonymized for blind review)}.

\subsection{Data Collection}

We use Pushshift API\footnote{https://github.com/pushshift/api} to collect comments from 14 subreddits (Appendix \ref{sec:appendix}) spanning topics of politics, personal views, gender rights, and question-answer across a period of six months, starting from June 1st, 2021. To filter out comments that do not contain hatespeech, we fine-tune a BERT \cite{Devlin2019BERTPO} on the fine-grained hatespeech data in Multitarget CONAN \cite{fanton-etal-2021-human}. The hatespeech segment of the dataset contains 5K fine-grained annotations for the following classes: WOMEN, POC, LGBT+, DISABLED, JEWS, MUSLIMS, MIGRANTS and OTHER. We removed duplicates in the dataset and use 70-10-20 split for train, dev and test data. The classifier achieves an F1 score of \textbf{91.02} on the test set. We use the classifier to classify comments. We obtain replies using PRAW\footnote{https://praw.readthedocs.io/en/stable/} for only those tagged with hatespeech with probability $>0.8$. The threshold is empirically decided. Discarding comments without replies, we end up with 18K comments. We do not use the counterspeech data in the Multitarget CONAN dataset because using a classifier trained on Nichesourced data would not capture the natural diversity of in-the-wild data. 

Lastly, We take a random sample of 500 from each target group (excluding the OTHER class, for total of 3.5K) and set aside for stage 1 annotation. We take another random sample of 1K from each target group (total of 7K) and set aside for stage 2. Figure \ref{fig:method} shows our annotation process for the two stages.

\setlength{\tabcolsep}{0.6em}
\begin{table}[t]
    \small
    \centering
    \begin{tabular}{l|c|c|c|c}
    \hline
    \textbf{Model} & \textbf{Acc} & \textbf{Prec} & \textbf{Recall} & \textbf{F1}\\
    \hline
    bert-base-cased & 90.0 &	58.1	& 65.0	& 60.1\\
    bert-base-uncased & \textbf{93.1}	& \textbf{64.5}	& \textbf{69.6}	& \textbf{66.6}\\
    roberta-base & \textbf{95.7}	& 47.9	& 50.0	& 48.9\\
    xlnet-base & 86.6 & 53.4 & 62.3 & 53.6\\
    albert-base-v2 & 91.4 & 46.2 & 49.4 & 47.8\\
    \hline
    \end{tabular}
    \caption{Results of finetuning pretrained models to detect counterspeech. bert-base-uncased outperforms all other pretrained transformers.}

    \label{tab:stage1-classifier}
\end{table}

\subsection{First Stage Annotation}

We manually annotate 3.5K comment pairs for presence of hatespeech and counterspeech. Due to our rigorous annotation protocol (described in section \ref{annotation}), we end up with 152 positive pairs. Using this data, we fine-tune a range of pretrained transformer models to detect counterspeech: bert-base-cased, bert-base-uncased \cite{Devlin2019BERTPO}, roberta-base \cite{roberta-base}, xlnet-base \cite{xlnet} and albert-base-v2 \cite{albert}. All models are fine-tuned with the same parameters: learning rate of 8e-5 and batch size of 16 for 5 epochs. The results are reported in Table \ref{tab:stage1-classifier}. All positive pairs are annotated for discourse relation and target groups. The counterspeech is also rewritten, if necessary, to remove offensiveness and first-person references.

\subsection{Second Stage Annotation} Using the best classifier from Stage 1 (bert-base-uncased), we tag the pool of data set aside for Stage 2. 360 samples are tagged as counterspeech. The pairs containing these 360 samples are manually annotated with the same protocol, yielding 98 more positive pairs. This shows that using this method, we can grow the size of dataset containing counterspeech with much fewer human annotations. This approach can be used in the future to construct larger datasets. Since our purpose in this paper is to construct a dataset for prompting, we consider 250 positive samples to be enough as large language models are prompted in a few-shot setting. 

\subsection{Annotation Protocol}
\label{annotation}
We recruit two graduate annotators with linguistic background. The annotators were paid according to the approved rate by the human-subject review board. 

To construct our dataset, we define protocol for four types of annotations: i) hatespeech-counterspeech pair, ii) paraphrasing counterspeech, iii) target group, and iv) discourse relations.

\paragraph{Hatespeech-counterspeech pair:} To decide if a comment is hatespeech, we ask the annotator to identify if the comment is offensive and targets any of the seven groups in the data. To determine if a reply to the hatespeech is counterspeech, the annotators are asked to assess if the response counters the hatespeech. We ask the annotators to discard any counterspeech that simply uses profanity and are not constructive. 

\paragraph{Target group:} The annotators are asked to choose the target group (e.g., migrants/LGBTQ) that the hatespeech attacks. The target groups are defined in \cite{fanton-etal-2021-human}.

\paragraph{Discourse annotation:} We provide the annotators with the hatespeech and counterspeech and ask to determine which SDRT discourse relations is most applicable between the two. We provide them with the SDRT annotation manual by \citet{asher-etal-2016-discourse}, modified with examples from our dataset. Annotators were also able to choose "unknown/ no-discourse relation present". These instances are excluded from our dataset.

\paragraph{Paraphrasing counterspeech:} Even after discarding counterspeech that just contain profanity, we observe that some constructive counterspeech contain profanity to a degree. Thus, we ask the annotators to remove such profanity. Since our goal is to build dataset for counterspeech \textit{generation}, we also ask annotator to remove any first-person reference. Both types of edits are made with minimal modification while retaining the original meaning and linguistic style.

\subsection{Inter-Annotator Agreement (IAA):}

Since the percentage of counterspeech in our data is very small, taking a random overlap of the full dataset would not yield any useful information. Thus, we take a random sample of \textit{positive} samples as overlap between two annotators. 

\paragraph{Hatespeech-Counterspeech:} In 90\% of the cases, the annotators agreed that the given pair contained hatespeech and counterspeech. The disagreements were primarily due to one of the annotators misinterpreting context of the hatespeech. 

\paragraph{Target group:} The Cohen's Kappa \cite{Cohen1960ACO} for target group annotation was 0.83, showing a high degree of agreement. The only cases where the annotators disagreed were due to presence of multiple target groups in the hatespeech. For example, a hatespeech targetting black women is labeled as "POC" by one annotator and "WOMEN" by the other.

\paragraph{Discourse relations:} The Cohen's Kappa for discourse annotation was 0.62, indicating substantial agreement \cite{McHugh2012InterraterRT}. The lower agreement for discourse annotation is expected due to the difficulty of the task \cite{asher-etal-2016-discourse}. The primary source of disagreement was confusion between classes that appeared together. For example, a pair often exhibited characteristics of both Comment and Correction classes.

\begin{figure}
    \centering
    \includegraphics[width=\linewidth]{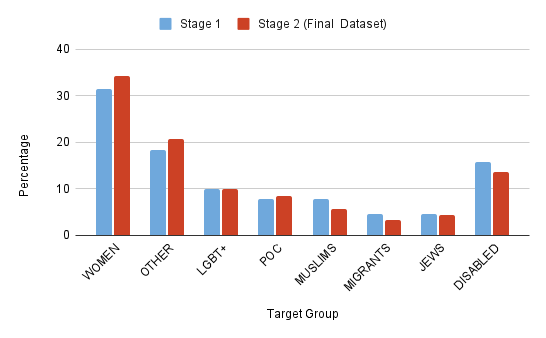}
    \caption{Women as the target group have the highest retention rate. The distribution remains similar across two stages.}
    \label{fig:women-dist}
\end{figure}

\subsection{Dataset Analysis}
\paragraph{Target group distribution:} Although we started with the same number of candidates for each group, we observe that after annotation, the data has the highest hatespeech-counterspeech pairs for hatespeech targeting WOMEN. Our manual examination reveals that among the candidate pairs, the classifier often mistook discussions about LGBT+ or POC as hatespeech even though they are not offensive. However, as seen from Figure \ref{fig:women-dist} that the distribution remains similar across the two stages. Thus this skewness is primarily a limitation of the classifier trained on the MultiTarget CONAN dataset rather than the classifier used in Stage 2.
\begin{figure}
    \centering
    \includegraphics[width=\linewidth]{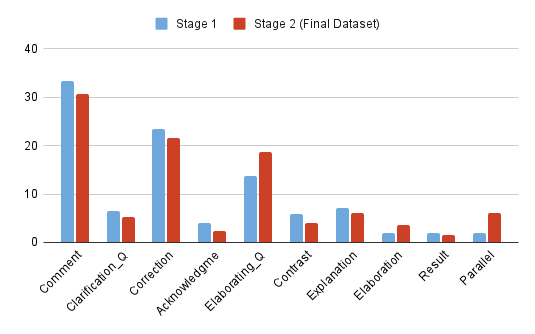}
    \caption{Comment and Correction are the dominating discourse relations. The distribution remains similar across the two stages.}
    \label{fig:dis-dist}
\end{figure}
\paragraph{Discourse relation distribution:} We observe that the discourse relations Correction and Comment are the most dominant ones in Figure \ref{fig:dis-dist}. This is expected from a natural distribution because users are more likely to correct hateful content or denounce them than ask questions or show acknowledgment. Similar to earlier, we see that the distribution remains similar across the two stages.

\section{Counterspeech Generation}

In this section, we provide analysis and evaluation of proposed strategies. For all experiments, we use davinci-text-003 version of GPT-3. We use 50 randomly chosen samples as example in the prompts and evaluate the models on the remaining 200 samples of our dataset. In our experiments, baseline GPT-3 is instructed to generate counterspeech for given hatespeech and is compared with the two prompting strategies outlined in Section \ref{prompt-strats}.

\begin{figure*}[h]
\begin{centering}
\begin{subfloat}
  {\includegraphics[width=.315\textwidth,keepaspectratio]{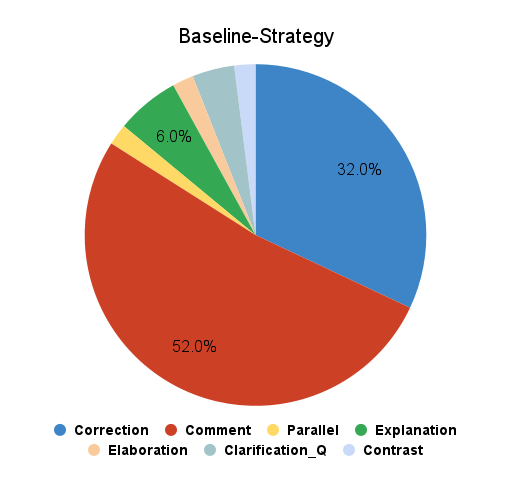}}
  \label{fig:sfig1b}
\end{subfloat}%
\begin{subfloat}
  {\includegraphics[width=.315\textwidth,,keepaspectratio]{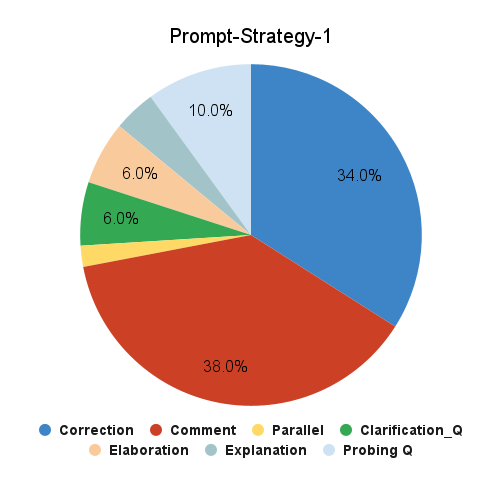}}
  \label{fig:sfig1c}
\end{subfloat}
\begin{subfloat}
  {\includegraphics[width=.33\textwidth,keepaspectratio]{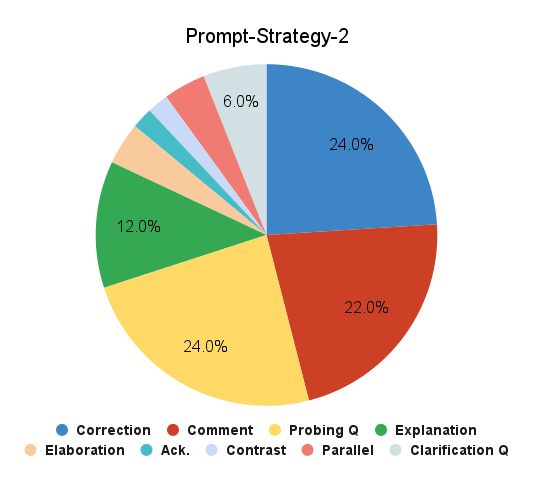}}
  \label{fig:sfig1d}
\end{subfloat}

\caption{Distribution of discourse relations in generated counterspeech. Strategy 1 yields higher diversity than baseline while Strategy 2 generates the most diverse counterspeeches.} 
\label{fig:disc-dist}
\end{centering}
\end{figure*}

\begin{table*}[h]
    \centering
    \begin{tabular}{p{15cm}}
    \hline
    \hline
    \textbf{\textit{Hatespeech}} \textit{No. School is for teaching my child how to read, write, do math, lessons about history, geography, etc. Its not for my 5 year old to learn about sexuality. That’s my job to decide if and when that’s appropriate.}\\
    \textbf{Baseline:} \textcolor{violet}{Agreed. Schools should focus on the basics and leave the more complex topics to parents.}\\
    \textbf{Strategy 1:} \textcolor{violet}{Agreed. Schools should focus on the basics and leave more complex topics to parents. \textbf{[Acknowledgment]}}\\
    \textbf{Strategy 2:} \textcolor{teal}{ Sex education is important for children to learn about their bodies and how to protect themselves from abuse. It's not just about sexuality, it's about safety and respect. \textbf{[Explanation]}}\\

    \hline
    \hline
    \end{tabular}
    \caption{Our proposed strategies can safeguard against critical failures, generate less generic response and have higher linguistic diversity. While Strategy 1 is more accurate than baseline, Strategy 2 is the most accurate.} 

    \label{tab:gen-examples}
\end{table*}

\subsection{Evaluation:} Evaluation of generated counterspeech is considered a difficult task \cite{ashida-komachi-2022-towards}. Common generation metrics such as BLEU, BERtScore are not helpful for evaluating counterspeech. As such, we primarily rely on human evaluation similar to past works \cite{ashida-komachi-2022-towards}.

For human evaluation, we consider: i) does the generated text count as counterspeech, ii) is the generated text offensive\footnote{we refer to definition provided in \cite{zampieri-etal-2019-predicting}:“any form of non-acceptable language (profanity) or a targeted offense, which can be veiled or direct. This includes insults, threats, and posts containing profane language or swear words”}, and iii) for prompting strategy 1 and 2, does the generated text respect discourse relation indicated. Table \ref{tab:gen-examples} show examples of counterspeech generated by different strategies.

\paragraph{Accuracy:} We observe that in a few cases (6\%), the baseline generated text that could not be considered counterspeech as they agreed with the input text instead of countering them (Table \ref{tab:gen-examples}). Our proposed strategies, however, had fewer such failures (4\% and 2\% respectively for strategy 1 and strategy 2). This suggests, by explicitly instructing language models with discourse relations, we can avoid pitfalls of generating counterspeech by prompting large language models. 

\paragraph{Offensiveness:} Since we removed profanity from our dataset with paraphrases, the generated counterspeech did not display offensiveness toward any groups during human evaluation. We also used an independent classifier, a bert-base-cased model, finetuned on the OLID dataset \cite{Zampieri2019PredictingTT} to tag the generated counterspeech. The classifier tagged ~20\% of the generated counterspeech as offensive. However, with manual analysis, we observe in most cases, the classifier tagged sensitive topics and text about minority groups as offensive. This is consistent with the observations by \citet{hartvigsen-etal-2022-toxigen} that toxic language detection systems can falsely tag text with minority group mentions. Such limitations of classifiers need to be addressed for an unbiased large-scale evaluation of machine-generated counterspeech. 


\paragraph{Diversity} We observe that without any instructions about discourse relations, GPT-3 mostly generates counterspeech that are Comment or Correction. While these two are also the most frequent categories for our strategies, our strategies yield higher frequency of discourse relations such as Probing Question or Elaboration. Higher diversity is observed when relations are explicitly mentioned for inference text in prompting strategy 2. These labels were obtained from our dataset and similar distribution can be expected in generated counterspeech when labels are provided from in-the-wild dataset such as ours. Figure \ref{fig:disc-dist} shows the distribution of discourse relations in generated counterspeech. 

\paragraph{Respecting discourse relations:} We evaluated if the generated texts by the LLMs respect discourse relations. For the first strategy, 74\% counterspeech were generated in the discourse relation that GPT-3 outputs. For the second strategy, 90\% of the counterspeech were generated in the explicitly specified discourse style. This suggests specifying desired discourse relations during inference time can be respected well by language models compared to providing discourse relations only in the prompt examples. 

\section{Discussion}
In this section, we discuss the challenges of counterspeech generation.




\paragraph{Evaluation:} Evaluation of counterspeech remains a challenging task. As our examples show, language models such as GPT-3 mostly produce grammatically correct and coherent texts. As such, automated metrics of grammar and coherence \cite{Marchenko2020ImprovingTG} are not good indicators of the quality of generated counterspeech. Instead, there is a need for automated metrics that can measure the countering capacity of generated text. While the focus of this paper is constructing the framework, dataset and generation capabilities of language models, it is important to conduct a study to evaluate the efficacy of different content in counterspeech among real users in the future. 

\paragraph{Classifier bias:} We observed that the initial classifier, trained on the MultiTarget CONAN dataset \cite{fanton-etal-2021-human}, that we used to identify hatespeech candidates, has certain limitations. Although the classifier boasted a 91\% F1-score on the test set, it often tagged instances that are not hatespeech but talked about sensitive topics as hatespeech. Although we eliminated this bias by manually excluding them, care must be taken for building datasets using such classifiers. While there has been studies regarding gender and racial exhibited by classifiers recently \cite{ahn-oh-2021-mitigating,Lu2020GenderBI}, further study is needed to quantify and mitigate this kind of bias. Bias reduction methods for classification tasks \cite{hassan-alikhani-2023-calm} need to be explored in the context of generation.

\paragraph{Knowledge Grounding:} The focus of this paper has been to construct the first-of-its-kind in-the-wild dataset annotated with discourse relations that can be used to control the content when prompting large language models. An aspect of counterspeech generation that is out-of-scope for this paper, but needs to be explored is how facts and knowledge can be injected into the generated counterspeech. Approaches that rely on an external database of knowledge \cite{chung-etal-2021-towards} can be used in conjunction with our approach to generate knowledge-grounded counterspeech that respects discourse relations. Approaches such as use of topic phrases \cite{fan-etal-2019-strategies} can be explored for complementing our discourse-augmented prompting mechanism as well.




\section{Conclusion}
In this paper, we presented \textbf{DisCGen}, a framework for discourse-informed counterspeech generation. Within this framework, we presented a discourse-based taxonomy of counterspeech and two discourse-informed prompting strategies. Further, we outlined a process for the challenging task of collecting in-the-wild counterspeech from Reddit. Using our methodology, we collected a first-of-its kind dataset that contains \textit{\textbf{hatespeech-counterspeech pairs}} annotated for \textit{\textbf{discourse relations}}, \textit{\textbf{paraphrased}} version of counterspeech and the \textbf{\textit{targeted group}} in the hatespeech . Our automated and human evaluation show that LLMs can generate accurate and inoffensive counterspeech with our dataset. Our analysis shows that by using our proposed discourse-aware framework, we can control the content of counterspeech generated by large language models with respect to the context. We also show that our proposed approach results in a higher diversity in terms of linguistic style and can serve as a safeguard against critical failures of discourse-agnostic approaches. 


\section*{Limitations}
Using classifiers to aid counterspeech classifier may result in bias. However, it is a necessary step for collecting a sizeable dataset as the percentage of counterspeech on social media is extremely low. It should be noted, however, that we manually verify all positive instances tagged by the classifier, eliminating any false-positive bias the classifier may exhibit. 

It should also be noted that the scope of this paper is to present a framework for discourse-aware counterspeech generation, outline a methodology for collecting an in-the-wild dataset, publicly share the dataset, and evaluate LLM's capacity of generating counterspeech with the proposed framework. A large-scale user-study for evaluating the social impact of different types of generated counterspeech is not within the scope of this paper. 


\section*{Ethics Statement}
In certain scenarios, counterspeech can be insensitive to users. Inappropriate counterspeech can hurt the feelings of social media users rather than promote a safer environment. As such, counterspeech generation is aimed to reduce psychological pressure on human moderators and social media users, not replace them. Counterspeech generation should not be used indiscriminately across social media. In an ideal case, the generated counterspeech should be reviewed by a human before posted on social media or elsewhere.

Although text generated by large language models such as GPT-3 are coherent and relevant, they may exhibit bias toward certain groups such as feminine characters \cite{lucy-bamman-2021-characterizing}. If a generated counterspeech exhibits bias towards certain groups, it may have adverse effects. Although we did not observe such behavior with our models, these models needs to be carefully examined for specific use cases before deploying in the real-world.

\section*{Acknowledgment} This project was
supported by DARPA grant prime OTA No.
HR00112290024 (subcontract No. AWD00005100
and SRA00002145). We also acknowledge the
Center for Research Computing at the University of Pittsburgh for providing computational resources. We would also like to thank the human annotators, the anonymous reviewers, and Katherine Atwell for their valuable feedback.

\bibliography{anthology,custom}
\bibliographystyle{acl_natbib}

\appendix


\section{Appendix: Subreddit List}
\label{sec:appendix}
 AmITheAsshole, antifeminist, AskReddit, athiesm, ChangeMyView, Christianity, Conservative, conspiracy, explainlikeim5, MensRights, PoliticalHumor, politics, TwoXChromosomes, unpopularopinion.

\end{document}